\PassOptionsToPackage{numbers}{natbib}
    \documentclass{article}

     \usepackage[final]{neurips_2025}

\usepackage[utf8]{inputenc} 
\usepackage[T1]{fontenc}    
\usepackage{url}            
\usepackage{booktabs}       
\usepackage{amsfonts}       
\usepackage{graphicx}       
\usepackage{nicefrac}       
\usepackage{microtype}      
\usepackage{xcolor}         
\usepackage{amsmath}
\usepackage{wrapfig}
\usepackage{multirow} 
\usepackage{comment} 
\usepackage{hyperref}

\title{Better Tokens for Better 3D: Advancing Vision-Language Modeling in 3D Medical Imaging}

\author{%
{\fontsize{11}{13}\selectfont Ibrahim Ethem Hamamci}$^{1,}$%
\thanks{Authors may list their names first in their CVs. \textbf{Correspondence: \texttt{ibrahim.hamamci@uzh.ch}}}%
\And
{\fontsize{11}{13}\selectfont Sezgin Er}$^{1,}$\footnotemark[1] \And
{\fontsize{11}{13}\selectfont Suprosanna Shit}$^{1,}$\footnotemark[1]  \And
{\fontsize{11}{13}\selectfont Hadrien Reynaud}$^{3}$ \And
{\fontsize{11}{13}\selectfont Dong Yang}$^{2}$ \And
{\fontsize{11}{13}\selectfont Pengfei Guo}$^{2}$ \And
{\fontsize{11}{13}\selectfont Marc Edgar}$^{2}$ \And
{\fontsize{11}{13}\selectfont Daguang Xu}$^{2}$ \And
{\fontsize{11}{13}\selectfont Bernhard Kainz}$^{3,4}$ \And
{\fontsize{11}{13}\selectfont Bjoern Menze}$^{1}$  \And 
\\[-1.2em]
{\normalsize
$^{1}$ University of Zurich \quad
$^{2}$ NVIDIA \quad
$^{3}$ Imperial College London \quad
$^{4}$ FAU Erlangen-Nürnberg
}
}

\begin{document}

\maketitle

\begin{abstract}

Recent progress in vision-language modeling for 3D medical imaging has been fueled by large-scale computed tomography (CT) corpora with paired free-text reports, stronger architectures, and powerful pretrained models. This has enabled applications such as automated report generation and text-conditioned 3D image synthesis.  
Yet, current approaches struggle with high-resolution, long-sequence volumes: contrastive pretraining often yields vision encoders that are misaligned with clinical language, and slice-wise tokenization blurs fine anatomy, reducing diagnostic performance on downstream tasks. We introduce \textbf{BTB3D} (\emph{Better Tokens for Better 3D}), a causal convolutional encoder-decoder that unifies 2D and 3D training and inference while producing compact, frequency-aware volumetric tokens. A three-stage training curriculum enables (i)~local reconstruction, (ii)~overlapping-window tiling, and (iii)~long-context decoder refinement, during which the model learns from short slice excerpts yet generalizes to scans exceeding $300$ slices without additional memory overhead. BTB3D sets a new state-of-the-art on two key tasks: it improves BLEU scores and increases clinical F1 by 40\% over CT2Rep, CT-CHAT, and Merlin for report generation; and it reduces FID by 75\% and halves FVD compared to GenerateCT and MedSyn for text-to-CT synthesis, producing anatomically consistent $512\times512\times241$ volumes. These results confirm that precise three-dimensional tokenization, rather than larger language backbones alone, is essential for scalable vision-language modeling in 3D medical imaging. The codebase is available at: \url{https://github.com/ibrahimethemhamamci/BTB3D}

\end{abstract}

\section{Introduction}

Three-dimensional (3D) medical images, such as CT, provide a rich volumetric view of anatomy, offering significantly more detail than 2D radiographs~\cite{mccollough2023milestones}. This makes them well-suited for vision-language models (VLMs), which can automate radiology report generation and enable text-guided volume synthesis for data augmentation, education, and planning~\cite{zhang2024challenges}. Recent progress in 3D VLMs, driven by public datasets with paired reports (especially in chest CT) and advances in vision encoders and large language models (LLMs), has opened new clinical possibilities~\cite{hamamci2024developing, hatamizadeh2022unetr, li2023llava, touvron2023llama, wang2024internvideo2}. However, these systems still face major challenges: 3D volumes often consist of hundreds of slices, introducing long-sequence modeling challenges~\cite{pati2023gandlf}. Moreover, the limited availability of paired 3D data prevents robust training~\cite{hwang2025ai}. As a result, 3D VLMs struggle to capture fine-grained clinical details and maintain spatial coherence when tasked with generating detailed reports or high-fidelity 3D volumes~\cite{hamamci2024ct2rep, blankemeier2024merlin, hamamci2024generatect}.

\textit{Why do 3D VLMs lag behind their 2D counterparts in report generation?} A typical pipeline combines a pretrained 3D vision encoder with an LLM for report generation~\cite{hamamci2024developing}. While LLMs have improved significantly through large-scale pretraining, the vision side remains a bottleneck~\cite{blankemeier2024merlin}. Most state-of-the-art methods rely on vision encoders pretrained with contrastive objectives~\cite{zhao2025clip, shuilarge}. However, this approach faces two critical limitations in the 3D medical context. First, contrastive learning assumes that only paired image-text samples are semantically aligned, while unpaired samples are unrelated~\cite{radford2021learning}. In radiology, this assumption often does not hold: multiple reports can describe the same conditions differently due to variations in language, clinical focus, or radiologist style~\cite{jeblick2024chatgpt}. Penalizing such unmatched pairs may therefore degrade the model’s understanding of medical semantics. Second, CLIP-style training requires large batch sizes for stability~\cite{tiu2022expert}. High-resolution 3D volumes are memory-intensive, often forcing trade-offs in batch size or model depth, resulting in underpowered vision encoders~\cite{shuilarge}. Prior work has noted that key findings (\emph{e.g.}, small nodules or subtle textures) may be lost when compressing an entire 3D scan into a single vector using weak models trained contrastively~\cite{hamamci2024developing}. Another major challenge is the lack of interoperability between 2D and 3D representations~\cite{hamamci2024developing}. Most 2D medical VLMs cannot directly process volumetric input and instead require projecting 3D scans into 2D slices~\cite{tanida2023interactive, tu2024towards}. This modality gap limits the transferability of pretrained 2D models to 3D, particularly important given the scarcity of 3D image-text datasets.

\textit{A similar bottleneck affects text-conditional 3D medical image generation.} Prior methods rely on encoder-decoder networks pretrained in a self-supervised manner to reconstruct volumes~\cite{hamamci2024generatect}. However, training an encoder-decoder capable of reconstructing long, high-resolution CT sequences remains an open challenge. Thus, existing methods adopt cascaded generation frameworks, resulting in spatial discontinuities and reduced realism, or rely on lightweight encoder-decoder architectures that fail to capture nuanced context~\cite{xu2024medsyn}. Consequently, current models struggle to generate anatomically consistent, high-quality 3D volumes from text. Hence, we need encoder-decoder networks for 3D VLMs that (1) scale to long volumetric sequences without losing critical details, (2) bridge the 2D/3D divide through unified representations, and (3) decode high-resolution 3D volumes from these representations. Improved encoders would enable more effective alignment between CT scans and textual descriptions for report generation, while better decoders enhance text-to-image generation.

To address these challenges, we introduce \textbf{BTB3D} (Better Tokens for Better 3D), a novel encoder-decoder framework that advances VLMs in 3D medical imaging through improved tokenization, reconstruction, and training strategies. Our core contribution is a \emph{causal convolutional encoder-decoder} that learns a compact sequence of volumetric \emph{tokens} for each CT. We adopt a quantized latent space for efficient 3D tokenization and reconstruction~\cite{zhang2023regularized}. Causal 3D convolutions enable the model to process scans slice by slice (analogous to a temporal sequence), allowing scalability to arbitrarily long scans and compatibility with pretrained 2D features, serving as a \emph{bridge between 2D and 3D modalities}. For downstream tasks, the proposed architecture can be combined with modules such as a transformer decoder for report generation or a diffusion model for volume synthesis. Crucially, we introduce a novel \textit{three-stage training strategy} that progressively adapts the encoder-decoder to longer input contexts, enabling robust training even under compute-limited conditions.

\section{Related works}
\paragraph{Radiology report generation from 3D medical images.}
The recent availability of 3D medical datasets paired with corresponding reports (such as CT-RATE) has enabled significant progress in this domain~\cite{hamamci2024developing}. The first framework for radiology report generation from 3D scans, CT2Rep, did not leverage any pretrained vision or language models~\cite{hamamci2024ct2rep}. Due to limited training data and the need to generate long, meaningful reports, CT2Rep employed a relational memory module to retrieve relevant past reports for coherent generation, inspired by prior work~\cite{chen2020generating}. Subsequent models, such as CT-CHAT and Merlin, improved upon this by incorporating pretrained components~\cite{hamamci2024developing, blankemeier2024merlin}. These approaches followed a contrastive strategy to align image and text features during vision encoder pretraining~\cite{radford2021learning}. A more recent model, fVLM, pretrained its 3D vision encoder, using a more clinically relevant contrastive learning approach~\cite{shuilarge}. However, we do not include fVLM in our comparisons, as neither its report generation codebase nor model weights have been available.

\paragraph{Text-conditional 3D medical image generation.}
The first framework for text-conditioned 3D volume generation, GenerateCT, introduced a cascaded pipeline~\cite{hamamci2024generatect}. Due to computational limitations, it trained a 3D encoder-decoder network on low-resolution CT scans, followed by 2D diffusion steps for upsampling~\cite{abu2024udpm}. While this cascaded approach enabled high-resolution and long-sequence generation, the volumes suffered from poor interslice consistency. MedSyn addressed this limitation with a 3D architecture designed to improve spatial coherence; however, its lightweight design reduced image fidelity~\cite{xu2024medsyn}. Other 3D CT scan generation methods, such as MAISI, exist, but we do not include them in our comparisons, as they are not designed for text-conditional generation~\cite{guo2025maisi}.

\begin{figure}[t!]               
  \centering
\includegraphics[width=\textwidth]{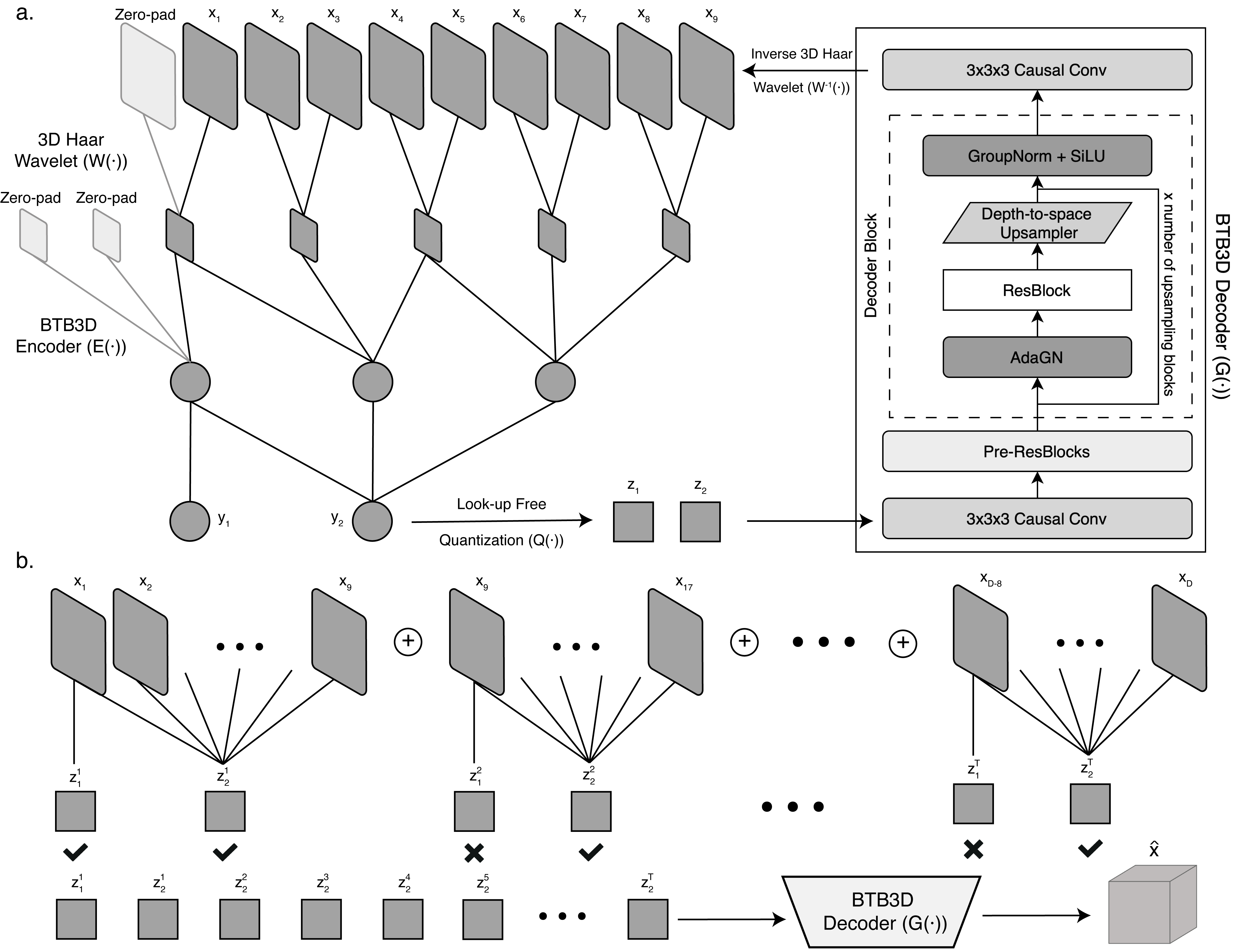}
 \caption{\textit{(a)} Stage~1: A 9-slice subvolume is compressed via a wavelet transform, then encoded causally using two stride-2 temporal convolutions. The encoder processes the input strictly causally by prepending zero-padded slices. Tokens are decoded back to the wavelet domain by a symmetric causal decoder. \textit{(b)} Stage~2: To scale to long CT volumes, we introduce overlapping temporal tiling (retaining only the second token from each window) to ensure consistent representation. Stage~3 follows the same scheme but trains only the decoder to refine long-range anatomical reconstruction.}
  \label{fig:fig1}
\end{figure}

\section{Methodology}
\label{section:3}

Due to the computational challenges, our goal is to pretrain on short sequences while enabling scalability to full volumes. Contrastive learning is inappropriate because paired reports typically describe the entire scan, and training on cropped or partial slices introduces semantic misalignment~\cite{zhao2025clip}. Also, they are difficult to adapt for 3D image synthesis~\cite{zeng2024make}. Thus, we propose a reconstruction-based pretraining, generalizing effectively to longer sequences. Prior work has shown that convolutional networks outperform transformer-based ones in video reconstruction~\cite{magvitv2}. Thus, we adopt a 3D CNN-based backbone~\cite{krizhevsky2012imagenet}. Given the limited 3D medical datasets, we design our model to support both 2D and 3D, enabling unified training across modalities, which motivates our causal network~\cite{liu2022structural}. 

However, training a convolutional model on short sequences introduces two key challenges: (1) When performing inference in a single pass over the full CT, reconstruction quality tends to degrade toward the latter slices; or (2) When performing inference in short chunks, inconsistencies arise at window boundaries. To address them, we introduce a three-stage training strategy that progressively scales while preserving temporal consistency. The resulting model can reconstruct long sequences with high fidelity and generalizes well across downstream tasks. Below, we detail the architecture and training.

\subsection{Model architecture}

BTB3D is a 3D convolutional encoder-decoder network with a discrete latent codebook~\cite{scannelldiscrete}. We design it for CT scans using causal convolutional mechanisms along the temporal (axial) dimension and wavelet-based compression for efficient and scalable input representation, shown in Figure~\hyperref[fig:fig1]{1a}. 


\paragraph{3D Haar wavelet compression.}
Given a volumetric CT scan \( x \in \mathbb{R}^{D \times H \times W} \), we apply a 3D Haar wavelet transform \( W(\cdot) \) to obtain a multi-channel representation \( W(x) \in \mathbb{R}^{\frac{D}{2} \times \frac{H}{2} \times \frac{W}{2} \times 8} \), following prior work in video generation that showed wavelet decompositions improve reconstruction quality~\cite{agarwal2025cosmos, tseng2024qtip, yao2022wave}. This transformation reduces resolution by a factor of two along each axis while retaining essential frequency information. Each non-overlapping \( 2 \times 2 \times 2 \) voxel block yields 8 subband coefficients: one low-frequency and seven high-frequency components. These frequency-aware channels encode both coarse and fine anatomical features, improving representation learning. Although the channel dimension increases from 1 to 8, the size of volumes is reduced by a factor of 8, yielding substantial compression for the input. This reduces memory and computation costs, enabling deeper or wider networks and making the model scalable to long, high-resolution CT volumes. For 2D slices (\( D=1 \)), the Haar transform along the \( z \)-axis produces a low- and high-frequency pair, with the latter often near-zero due to the absence of temporal variation. Nonetheless, the same 3D transform applies to both 2D and 3D inputs, ensuring architectural consistency across training modes.

\paragraph{Encoder with causal 3D convolutions.}

The encoder \( E \) takes a wavelet-transformed CT sequence \( W(x) \in \mathbb{R}^{D \times H \times W \times C} \) and outputs a latent representation \( y \in \mathbb{R}^{D' \times H' \times W' \times d} \). It consists of residual blocks with \textit{factorized 3D convolutions} that decouple spatial and temporal processing. Each block applies a \( 1 \times k \times k \) spatial convolution (sagittal-coronal) followed by a \( k \times 1 \times 1 \) temporal convolution (axial), with causal padding of \( (k{-}1) \) zero slices in the past and none in the future. This ensures that the encoder at index \( t \) only attends to slices \( \leq t \), enabling strictly causal encoding. As a result, the first token is computed from the first slice alone, and future leakage is prevented in all subsequent tokens. This causal design supports unified 2D (e.g., single-slice) and 3D (volume-level) training while preserving axial consistency, beneficial for downstream autoregressive tasks~\cite{kalchbrenner2016video, weissenborn2019scaling, chen2022simple}.

Figure~\hyperref[fig:fig1]{1a} illustrates the model architecture and causal axial compression. Downsampling is performed using strided convolutions interleaved with residual blocks. In the \textit{8$\times$8$\times$8} configuration, two stride-2 convolutions along each axis yield a \( 4\times \) reduction spatially and temporally. Combined with the initial \( 2\times \) wavelet downsampling~\cite{tseng2024qtip}, this results in an effective \( 8\times \) compression. The \textit{16$\times$16$\times$8} variant adds an extra spatial stride-2 layer, achieving \( 16\times \) spatial and \( 8\times \) temporal compression. Each variant presents a trade-off: the \( 8^3 \) setting preserves more spatial detail for tasks like segmentation or volume synthesis, while the \( 16^2{\times}8 \) variant offers higher compression for memory-constrained settings or tasks focused on global semantics, such as classification or radiology report generation.


\paragraph{Lookup-free quantization and decoder.}
The encoder output \( y = E(W(x)) \in \mathbb{R}^{D' \times H' \times W' \times d} \) is transformed into a discrete latent representation using \textit{lookup-free quantization}~\cite{tseng2024qtip}. The number of quantization codes is \( K = 2^d \), and each feature vector at a spatial position is binarized independently across dimensions by computing the sign of each element, resulting in binary vectors \( b \in \{-1, 1\}^d \). These vectors are then packed into integers, forming the token map \( z \in \{0, \dots, K-1\}^{D' \times H' \times W'} \). This approach removes the need for a codebook or embedding lookup during training and inference, significantly improving speed and memory efficiency, critical for large 3D volumes. To prevent code collapse (\emph{e.g.}, overuse of a limited set of tokens), we include an entropy regularization term \( \mathcal{L}_{\text{entropy}} \) that encourages uniform usage of all \( K \) codes. Prior work in discrete representation learning~\cite{esser2021taming, magvitv2} has shown this regularization improves both reconstruction and token diversity. The decoder \( G \), which mirrors the encoder’s structure as shown in Figure~\hyperref[fig:fig1]{1a}, employs transposed convolutions and residual upsampling blocks to reconstruct the wavelet-domain volume \( \tilde{W}(x) \). Like the encoder, the decoder is also \textit{causal}: each slice is reconstructed using only previously decoded slices, preserving autoregressivity~\cite{ramesh2022hierarchical}. The decoder accepts either the packed integer tokens or their binary expansions. The final volume \( \hat{x} \) is obtained by applying \( W^{-1} \), the inverse 3D Haar wavelet transform, to \( \tilde{W}(x) \).

\paragraph{Training objectives.}
The model is trained to reconstruct high-resolution 3D CT sequences using a combination of three objectives: reconstruction loss, adversarial loss, and quantization loss.

The \textit{reconstruction loss} \( \mathcal{L}_{\text{rec}} \) encourages the decoder to accurately recover the input volume:
\begin{align}
\mathcal{L}_{\text{rec}} 
= \mathbb{E}_{x} \left[ \left\| x - \hat{x} \right\|_1 \right] 
= \frac{1}{N} \sum_{i=1}^{N} \left| x_i - \hat{x}_i \right|,
\end{align}
where \( \hat{x} \) is the reconstructed volume and \( N \) is the number of voxels. We use the \( \ell_1 \) norm, which promotes sharper reconstructions and preserves fine anatomical detail better than \( \ell_2 \)~\cite{zhao2016loss}. 

The \textit{adversarial loss} \( \mathcal{L}_{\text{adv}} \) improves perceptual realism by encouraging indistinguishability:
\begin{align}
\mathcal{L}_{\text{adv}} 
= \mathbb{E}_{x} \left[ -\log D(\hat{x}) \right] 
= -\frac{1}{N} \sum_{i=1}^{N} \log D(\hat{x}_i),
\end{align}
where \( D \) is a 3D discriminator trained to distinguish real from generated CT volumes. We apply this supervision directly in the CT domain rather than the wavelet domain, stabilizing training~\cite{goodfellow2014generative, esser2021taming}.

The \textit{quantization loss} \( \mathcal{L}_{\text{vq}} \) enforces commitment to discrete representations:
\begin{align}
\mathcal{L}_{\text{vq}} = 
\mathbb{E} \left[
\left\| \text{sg}[y] - e \right\|_2^2 
+ \beta \left\| y - \text{sg}[e] \right\|_2^2
\right],
\end{align}
where \( e \) is the discrete embedding (from a codebook or binary quantization), and \( \text{sg}[\cdot] \) denotes the stop-gradient operator~\cite{van2017neural}. The overall training objective is a weighted sum:
\begin{align}
\mathcal{L} = \mathcal{L}_{\text{rec}} + \lambda_{\text{adv}} \mathcal{L}_{\text{adv}} + \mathcal{L}_{\text{vq}},
\end{align}
where \( \lambda_{\text{adv}} \) controls the influence of adversarial supervision. We omit perceptual losses (\emph{e.g.}, VGG-based features) due to the mismatch between natural RGB images and grayscale medical ones~\cite{ledig2017photo, zhang2018perceptual}. Prior work has shown that these losses can degrade performance in medical image reconstruction~\cite{zhou2020hi}.

\subsection{Three-stage training}

\paragraph{Stage 1: Short-volume pretraining.} We first train the model on single 2D slices or short 9-slice subvolumes. The entire model (encoder, quantizer, and decoder) is optimized end-to-end:
\begin{equation}
\min_{E,G} \mathcal{L}(x_{1:9}, \hat{x}_{1:9}), \quad 
\text{where} \quad \hat{x}_{1:9} = W^{-1}(G(Q(E(W(x_{1:9}))))).
\end{equation}
This phase helps the model learn local spatial and spatio-temporal structures.
\paragraph{Stage 2: Overlapping temporal tiling.} \label{stage2}We continue training both the encoder and decoder using overlapping short subsequences instead of full-volume encoding. To scale to long volumes while preserving temporal consistency, we adopt an overlapping window strategy. Specifically:
\begin{enumerate}
  \item Encode $x_{1:9} \Rightarrow z_1^1, z_2^1$. Keep both tokens.
  \item Encode $x_{9:17} \Rightarrow z_1^2, z_2^2$. Discard $z_1^2$ (covers only slice 9); keep $z_2^2$ (covers $x_9$–$x_{17}$).
  \item Encode $x_{17:25} \Rightarrow z_1^3, z_2^3$. Discard $z_1^3$; keep $z_2^3$. Repeat this pattern until the end of the CT.
\end{enumerate}
In each 9-slice window, we discard the first token and retain the second, which encodes information from all 9 slices (except the first window in which we take both tokens). This overlapping strategy promotes temporal consistency. Letting $z_i^t$ denote the $i$-th token from the $t$-th window, we have:
\begin{align}
\text{1st window: } [z_1^1, z_2^1] &= E(W(x_{1:9})) \\
\text{2nd window: } [z_1^2, z_2^2] &= E(W(x_{9:17})) \Rightarrow \text{keep only } z_2^2 \\
&\dots \nonumber \\
\text{$T$-th window: } [z_1^T, z_2^T] &= E(W(x_{(D-8):D})) \Rightarrow \text{keep only } z_2^T
\end{align}
where $T = \lfloor (D - 1)/8 \rfloor$, and $D$ is the sequence length of the partial CT volume that fits into memory in this second stage training. As shown in Figure~\hyperref[fig:fig1]{1b}, the final latent sequence is:
\begin{align}
[z_1^1, z_2^1, z_2^2, z_2^3, z_2^4, z_2^5, \dots, z_2^T].
\label{tokens}
\end{align}
This allows for efficient training on longer sequences than Stage 1, thanks to its lower memory footprint compared to one-shot encoding. The decoder processes all tokens in a single forward pass.

\paragraph{Stage 3: Long-sequence decoder fine-tuning.}
This stage mimics Stage 2, but we freeze the encoder \( E \) and the codebook, and fine-tune only the decoder \( G \). The training objective is:
\begin{equation}
\min_{\theta_G} \; \mathbb{E}_{x_{1:D}} \left[ \mathcal{L}\left(x_{1:D}, \hat{x}_{1:D} \right) \right] \quad \text{subject to } E \text{ and codebook frozen},
\end{equation}
where \( \theta_G \) are the parameters of the decoder \( G \). This step enhances the decoder’s capacity to model long-range anatomical dependencies by training it to reconstruct CT without modifying the encoder. 


\paragraph{Inference.} 
BTB3D supports two inference strategies for reconstructing full-length 3D CT volumes: \emph{one-shot} and \emph{tiled} inference, offering flexibility based on memory constraints and desired consistency. In \emph{one-shot inference}, the entire volume \( x \) is encoded and decoded in a single pass:
\begin{align}
\hat{x} = W^{-1}(G(E(W(x)))).
\end{align}
This is efficient when memory allows full-volume processing but it may degrade on long sequences. In \emph{tiled inference}, we follow the overlapping tokenization scheme from \hyperref[stage2]{Stage~2} to ensure spatial coherence over long sequences. The input volume \( x \) is split into overlapping 9-slice windows, and:
\begin{align}
[z_1^t, z_2^t] = E(W(x_{s_t:s_t+8})), \quad 
\hat{x} = W^{-1}(G([z_1^1, z_2^1, z_2^2, z_2^3, \dots, z_2^T])).
\end{align}
This strategy mirrors the training and ensures consistent reconstructions with lower memory use. We adopt tiled inference for all downstream experiments due to its robustness and improved consistency.

\section{Experiments}
\label{section:4}

We evaluate the effectiveness of our model and three-stage training through reconstruction and two downstream tasks (report generation and text-to-CT synthesis). Our experiments address three key questions: (1) Does the three-stage training improve reconstruction over naive end-to-end training? (2) Can our tokenization enhance vision-language tasks like report generation? (3) How does BTB3D compare to state-of-the-art models in synthesizing high-resolution CT from clinical prompts?

\begin{wraptable}{r}{0.55\linewidth}
\vspace{-1cm}
\vspace{-\baselineskip}
  \caption{Reconstruction metrics after each stage of the three-stage training. We report full-volume reconstruction performance at each stage for two models with compression rates of 8$\times$8$\times$8 and 16$\times$16$\times$8.}
  \label{three_stage_table}
  \centering
  \renewcommand{\arraystretch}{1.1}
  \begin{tabular}{lcccc}
    \toprule
    \textbf{Stage} & \textbf{C. Rate} & \textbf{PSNR~$\uparrow$} & \textbf{SSIM~$\uparrow$} & \textbf{MSE~$\downarrow$} \\
    \midrule
    Stage 1 & $8^3$          & 9.350 & 0.206 & 0.117 \\
    Stage 2 & $8^3$          & 23.980 & 0.697 & 0.005 \\
    Stage 3 & $8^3$          & \textbf{28.166} & \textbf{0.760} & \textbf{0.001} \\
    \midrule
    Stage 1 & $16^2{\times}8$ & 11.067 & 0.353 & 0.079 \\
    Stage 2 & $16^2{\times}8$ & 23.808 & 0.700 & 0.005 \\
    Stage 3 & $16^2{\times}8$ & \textbf{26.750} & \textbf{0.749} & \textbf{0.002} \\
    \bottomrule
  \end{tabular}
  \vspace{-\baselineskip}
 \vspace{-0.5cm}
\end{wraptable}

\subsection{Three-stage training performance}

We assess BTB3D’s three-stage training via reconstruction metrics and ablations, demonstrating its scalability from short subvolumes to full-resolution 3D chest CTs. The strategy progressively (1) learns local spatial and short-range temporal features, (2) extends to longer sequences using overlapping tiling, and (3) enhances global coherence  through decoder-only fine-tuning. This enables strong representations for downstream tasks such as report and text-to-CT generation (Sections~\ref{report_g} and~\ref{ct_g}).

\paragraph{Experimental results.}

\begin{figure}[t!]
  \centering
  \includegraphics[width=\textwidth]{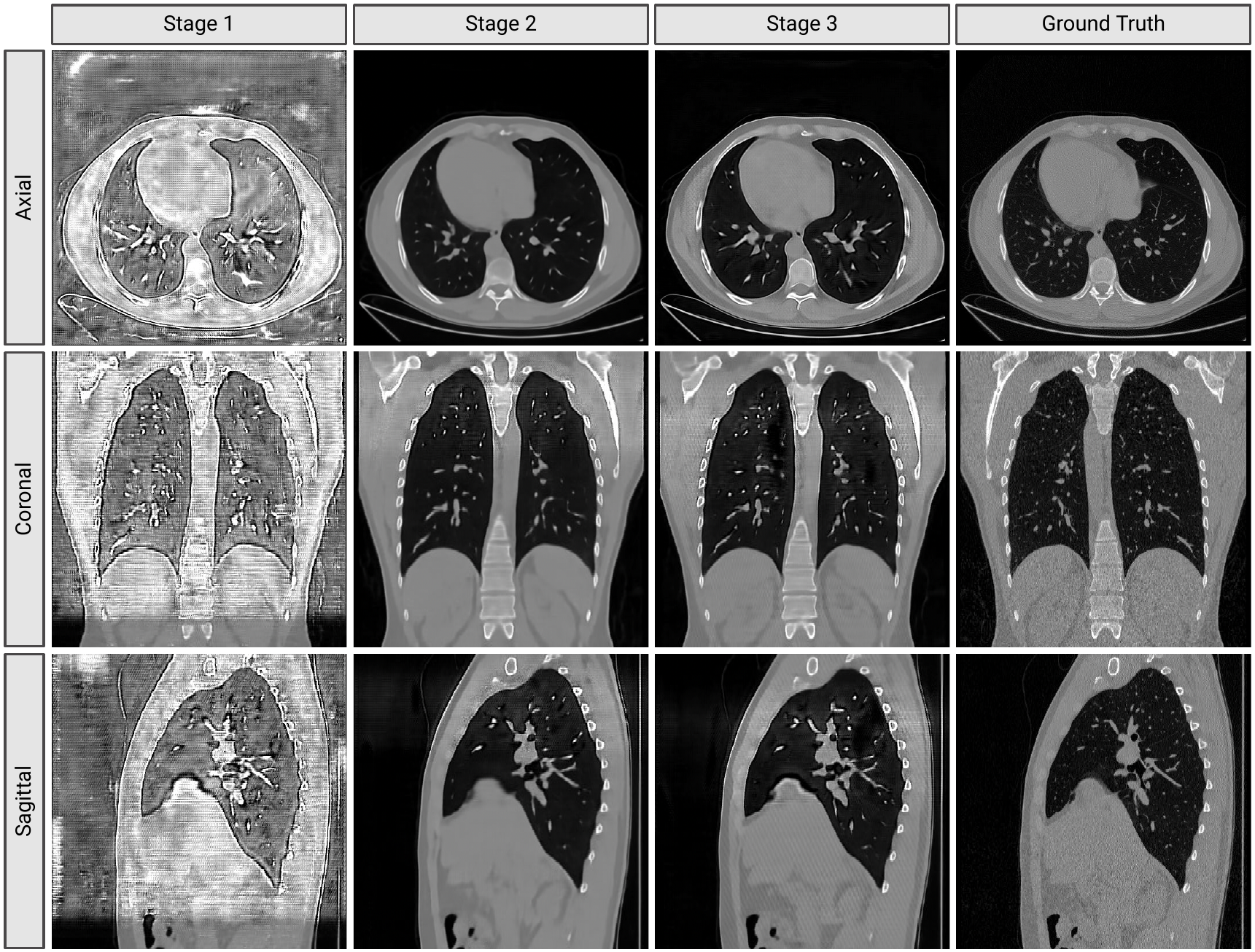}
  
   \caption{Reconstruction results of our 8$\times$8$\times$8 model across axial, coronal, and sagittal planes after each training stage. Progressive improvements demonstrate the effectiveness of our three-stage strategy, with Stage 2 providing the largest gains in anatomical fidelity and inter-slice consistency.}
   \label{fig:3stage}
\end{figure}

Table~\ref{three_stage_table} reports reconstruction performance across training stages~\cite{hore2010image, ssim}. Stage~1, trained on short subvolumes, captures local structure but fails at long-range consistency. Stage~2 introduces overlapping tiling, yielding the largest improvement: PSNR increases by over 14 dB, SSIM triples, and MSE drops an order of magnitude. Stage~3 refines inter-slice fidelity through decoder tuning, with modest but consistent gains. Figure~\ref{fig:3stage} visualizes reconstruction quality (8$\times$8$\times$8 variant) across planes. Stage~1 outputs are blurry and lose structure beyond 9 slices. Stage~2 restores coherent anatomy across distant slices. Stage~3 sharpens fine boundaries such as fissures and vessels.


\paragraph{Dataset and implementation.}
Aligned with the baselines, we use 25,692 chest CT scans from 21,304 patients in CT-RATE (the largest public 3D medical dataset with paired reports)~\cite{hamamci2024developing}. The training set includes 20,000 patients and the test set includes 1,304. Volumes are converted to Hounsfield Units (HU) and clipped to \([-1000, 1000]\)~\cite{denotter2019hounsfield, lamba2014ct}. For Stages~1 and~2, we use raw volumes; for Stage~3, volumes are resampled to \(0.75\times0.75\times1.5\) mm and cropped/padded to \(512\times512\times241\). Training is conducted on 64 NVIDIA H100 GPUs using DDP and mixed precision. In Stage~1, the batch size is 8 for 9-slice subvolumes and 40 for single slices. In Stages~2 and~3, we use a batch size of 1 (with 201 and 241 slices, respectively). The discriminator is reinitialized at each stage. We use Adam~\cite{diederik2014adam} with a learning rate of \(1\text{e}{-5}\), \(\beta_2 = 0.99\), and \(\epsilon = 1\text{e}{-8}\), along with gradient clipping (threshold: 0.5)~\cite{gradnorm}. LFQ uses a token dimension of 18 (codebook size: 262,144)~\cite{tseng2024qtip}. We train Stage~1 for 150k iterations, Stage~2 for 60k, and Stage~3 for 50k.

\subsection{Radiology report generation from 3D chest CT} \label{report_g}
\begin{wraptable}{r}{0.5\linewidth}
\vspace{-\baselineskip}
\vspace{-0.2cm}
  \caption{Evaluation on Rad-ChestCT proves strong out-of-distribution generalization. As only binary labels are available, text-based metrics are omitted.}
  \label{report-external-table}
  \centering
  \renewcommand{\arraystretch}{1.1}
  \begin{tabular}{lcccc}
    \toprule
    \textbf{Model} & \textbf{F1~$\uparrow$} & \textbf{Precison~$\uparrow$} & \textbf{Recall~$\uparrow$}  \\
    \midrule
    CT2Rep   & 0.133 & 0.299 & 0.139  \\
    Merlin   & 0.182 & 0.271 & 0.149  \\
    CT-CHAT  & 0.182 & \textbf{0.382} & 0.171  \\
    \midrule
    Ours-8  & 0.192 & 0.269 & 0.165  \\
    Ours-16   & \textbf{0.266} & 0.272 & \textbf{0.329}  \\
    \bottomrule
  \end{tabular}
  \vspace{-\baselineskip}
\end{wraptable}

We evaluate BTB3D on report generation for 3D CT. Each volume is encoded into a sequence of latent tokens \( \mathbf{v} \in \mathbb{R}^{T \times d_v} \) by the BTB3D encoder. For the $16 \times 16 \times 8$ model, we use 18-dimensional tokens; for the $8 \times 8 \times 8$ one, the token count is quadrupled and the embeddings are reduced to 72 via merging. A linear layer maps these tokens to the input space of LLaMA 3.1-8B~\cite{grattafiori2024llama}, which autoregressively generates reports following the CT-CHAT setup~\cite{hamamci2024developing}. We adopt the same configuration and pretrained LLM as CT-CHAT to ensure fair comparison and isolate the impact of our tokenization.

\paragraph{Baselines.}

Prior methods rely on lightweight encoders and contrastive pretraining, limiting tokenization quality to capture relevant details for report generation. In contrast, our pretraining (with architectural and training advances) produces richer representations. We benchmark BTB3D against three state-of-the-art report generation models. CT2Rep~\cite{hamamci2024ct2rep} uses a CT-ViT encoder and GPT-style decoder with memory modules. CT-CHAT~\cite{hamamci2024developing} aligns a CLIP-pretrained vision encoder with LLaMA~\cite{liu2023visual}. Merlin~\cite{blankemeier2024merlin} employs an I3D-ResNet backbone pretrained with masked and contrastive objectives and attaches a decoder similar to CT-CHAT. We use official weights for CT2Rep and CT-CHAT, both trained on CT-RATE. As Merlin was originally trained on private data and does not release weights, we retrain it on CT-RATE using the official codebase.

\paragraph{Experimental results.}

Table~\ref{report_internal_table} reports clinical accuracy (F1, precision, recall, CRG~\cite{hamamci2025crg}) and language quality (BLEU, METEOR~\cite{bleu, meteor}) on CT-RATE. Merlin achieves high precision but low recall, tending to under-report findings, while CT2Rep and CT-CHAT show high recall but lower precision, often hallucinating abnormalities. Our method achieves a better balance, reflected in superior F1 and CRG scores. The $16 \times 16 \times 8$ variant achieves the highest F1, with a 40\% relative improvement over CT-CHAT, confirming the effectiveness of our volumetric tokenization. Compared to CT-CHAT, BLEU-1 improves by 18\% and BLEU-mean by 12\%. Table~\ref{report-external-table} presents results on RadChestCT. Our $16 \times 16 \times 8$ model again achieves the highest F1 score, a 46\% relative improvement over the best baselines. Recall also improves substantially, indicating strong generalization to out-of-distribution data. All clinical metrics are computed using the official CT-RATE report classifier.

\paragraph{Dataset and implementation.}
We use the CT-RATE dataset for report generation training, with the findings and impression sections used as in prior baselines. Chest CT scans are resampled to a voxel spacing of \(0.75\times0.75\times1.5\) mm and padded or cropped to a uniform size of $512 \times 512 \times 241$. For external validation, we use RadChestCT~\cite{draelos2021machine}, which contains multi-label annotations but no text reports. Models are trained for 40,000 iterations using DeepSpeed ZeRO-3 on 40 NVIDIA H100 GPUs. We use the AdamW optimizer with a learning rate of \( 2 \times 10^{-5} \) and a warm-up ratio of 0.03. Each GPU processes one sample, yielding an effective global batch size of 40. For LoRA, we set \( r=64, \alpha=128 \) for the $8^3$ variant, and \( r=128, \alpha=256 \) for the $16^2{\times}8$ variant.

\begin{table}[t]

\caption{Report generation performance on CT-RATE. BTB3D outperforms prior methods in both metric types, with the higher compression variant achieving the highest F1 and BLEU scores.}

  \label{report_internal_table}
  \centering
  \renewcommand{\arraystretch}{1.1}
  \begin{tabular}{lccccccccccc}
    \toprule
    \multicolumn{1}{c}{} 
    & \multicolumn{4}{c}{\textbf{Clinical Accuracy~$\uparrow$}} 
    & \multicolumn{6}{c}{\textbf{Natural Language Generation~$\uparrow$}} \\
    \cmidrule(lr){2-5} \cmidrule(lr){6-11}
    Model & F1 & P & R & CRG & B$_1$ & B$_2$ & B$_3$ & B$_4$ & B$_\text{mean}$ & M \\
    \midrule
    CT2Rep   & 0.160 & 0.435 & 0.128 & 0.359 & 0.372 & 0.292 & 0.243 & 0.213 & 0.280 & 0.197 \\
    Merlin   & 0.160 & 0.295 & 0.112 & 0.352 & 0.231 & 0.163 & 0.124 & 0.099 & 0.154 & 0.148 \\
    CT-CHAT  & 0.184 & \textbf{0.450} & 0.158 & 0.368 & 0.373 & 0.284 & 0.231 & 0.198 & 0.272 & 0.215 \\
    \midrule
    Ours-8   & 0.187 & 0.260 & 0.150 & 0.357   & 0.411 & 0.307 & 0.245 & \textbf{0.215} & 0.295 & 0.220 \\
    Ours-16  & \textbf{0.258} & 0.260 & \textbf{0.260} & \textbf{0.370} & \textbf{0.439} & \textbf{0.320} & \textbf{0.248} & 0.213 & \textbf{0.305} & \textbf{0.223} \\
    \bottomrule
  \end{tabular}
\end{table}
\begin{table}[t]
\caption{Text-conditional CT generation results. BTB3D with lower compression outperforms previous methods across all metrics, showing superior consistency, image quality, and text alignment.}

  \label{ct_generation_table}
  \centering
  \renewcommand{\arraystretch}{1.1}
  \setlength{\tabcolsep}{5.65pt}  

  \begin{tabular}{lcccccccc}
    \toprule
    \multicolumn{1}{c}{} 
    & \multicolumn{4}{c}{\textbf{FID}~$\downarrow$} 
    & \multicolumn{2}{c}{\textbf{FVD}~$\downarrow$} 
    & \multicolumn{2}{c}{\textbf{CLIP Score}~$\uparrow$} \\
    \cmidrule(lr){2-5} \cmidrule(lr){6-7} \cmidrule(lr){8-9}
    Model & Axial & Sagittal & Coronal & Mean & CT-Net & I3D & Text-Img & Img-Img \\
    \midrule
    GenerateCT & 10.416 & 10.365 &  7.754 &  9.512 & 7.659 & 1512.5 & 23.625 & 84.287 \\
    MedSyn     & 14.963 & 12.115 & 10.698 & 12.592 & 13.927 & 725.81 & 23.571 & 84.153 \\
    \midrule
    Ours-8    & \textbf{2.479} &  \textbf{2.166} &  \textbf{2.062} &  \textbf{2.236} & \textbf{3.955} & \textbf{325.51} & \textbf{24.270} & \textbf{88.352} \\
    Ours-16     & 7.077 &  4.226 &  3.729 &  5.011 & 4.020 & 429.34 & 23.322 & 84.957 \\
    \bottomrule
  \end{tabular}
\end{table}

\subsection{Text-conditional 3D chest CT generation}\label{ct_g}

We evaluate BTB3D’s encoder-decoder on generating 3D chest CT scans from free-text prompts, assessing the realism, anatomical coherence, and text alignment of its latent representations. For generation, we use a transformer-based model with cross-attention layers, similar to GenerateCT~\cite{hamamci2024generatect}.


\begin{figure}[t!]
  \centering
  \includegraphics[width=\textwidth]{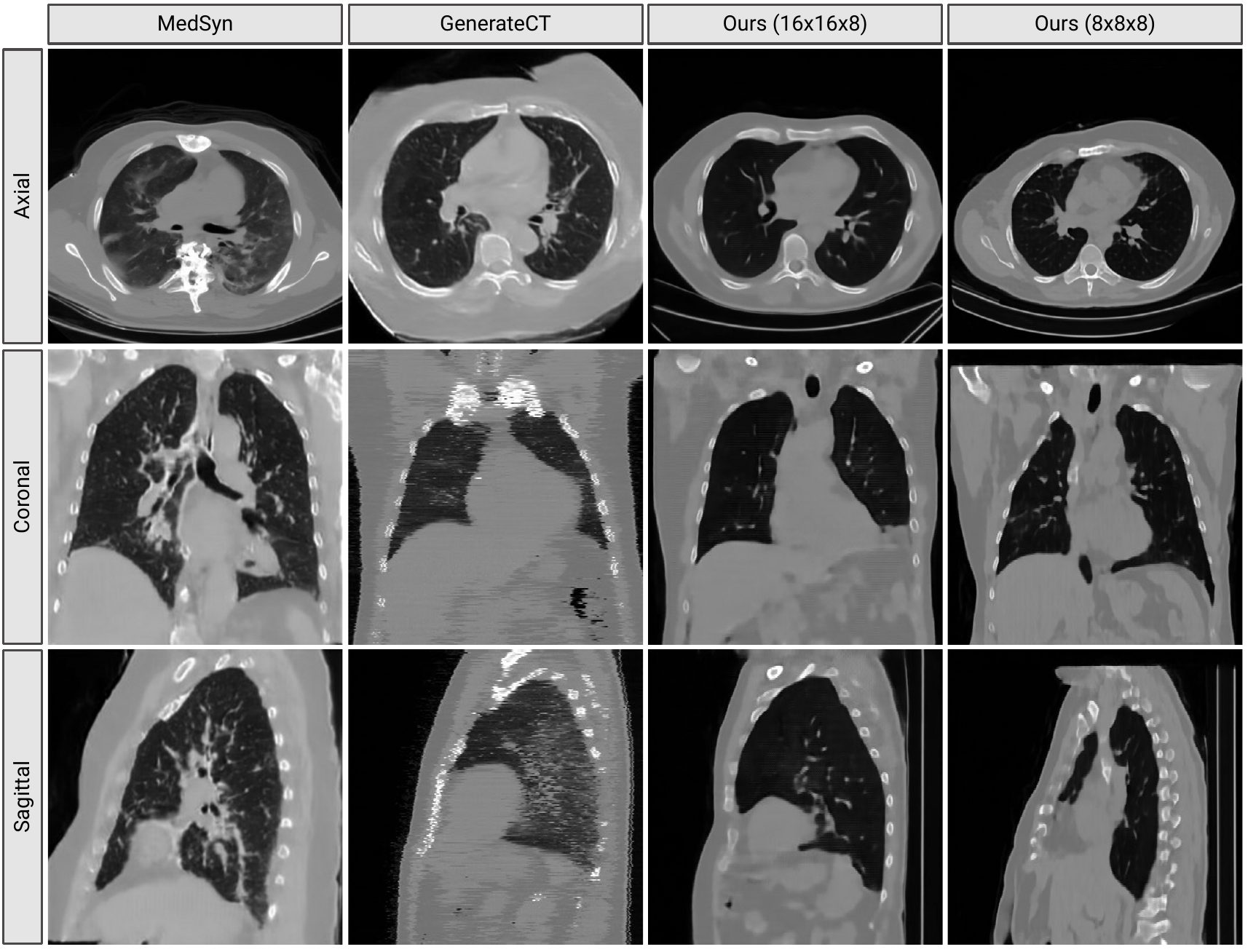}

  \caption{Example generations for a clinical prompt: \textit{“Chest CT scan of a 63-year-old male: No findings compatible with pneumonia were detected. Mild sequelae changes are observed in both lungs. Atherosclerotic changes noted, with slight increase in the calibration of vascular structures in the mediastinum. Hepatosteatosis. Hiatal hernia.”}. We show one representative slice per anatomical plane. Ground-truth volumes are omitted, following standard practice in text-to-image generation.}
\vspace{-0.19cm}
\label{ct_generation_fig}
\end{figure}

\paragraph{Baselines.}  GenerateCT~\cite{hamamci2024generatect} adopts a cascaded design (low-resolution 3D generation followed by 2D diffusion upsampling), which leads to spatial inconsistencies. This approach is motivated by the computational constraints of training a high-capacity 3D encoder-decoder directly on high-resolution volumes. MedSyn~\cite{xu2024medsyn} employs a unified 3D transformer with windowed attention to improve inter-slice consistency but suffers from lower fidelity due to its lightweight architecture.

\paragraph{Experimental results.}  
Table~\ref{ct_generation_table} reports FID (per view and mean), FVD (with CT-Net~\cite{draelos2021machine} and I3D backbones), and CLIP-based alignment scores. Since CT-Net is trained on 3D chest CT volumes, it provides a more relevant assessment than I3D. BTB3D with lower compression achieves substantial improvements: mean FID drops from 9.51 (GenerateCT) to 2.24 (a 76.5\% reduction). FVD$_\text{CT-Net}$ improves by 48.3\%, confirming better spatiotemporal realism, while CLIP alignment also slightly increases. The higher-compression BTB3D variant still outperforms GenerateCT and MedSyn across all metrics, though with smaller margins, highlighting a trade-off between compression and generation quality. These results underscore the effectiveness of our encoder-decoder network for high-fidelity, text-aligned 3D medical image synthesis. Figure~\ref{ct_generation_fig} illustrates qualitative differences: our model generates sharper volumes with clearer anatomical structures and better alignment to the clinical prompt, while MedSyn and GenerateCT outputs appear blurrier or contain inter-slice artifacts.

\paragraph{Dataset and implementation.}  
We use CT-RATE, with CTs resampled as in Section~\ref{report_g}. Prompts are generated as: \emph{"Chest CT scan of a \{age\}-year-old \{sex\}: \{impression\}"}, following GenerateCT. The generation model is a 12-layer transformer (1024 hidden size, 16 heads), trained using flow matching loss~\cite{ma2024sit}. We apply $[7,7,7]$ windowed self-attention and $[2,2,2]$ patching. Prompts are encoded using the T5v1.1-base model~\cite{raffel2020exploring}. Training is conducted on 16 H100 GPUs for 1500 epochs, using AdamW with a learning rate of $10^{-4}$. The batch size is 4 for the $8^3$ variant and 8 for $16^2{\times}8$.


\section{Conclusion}
We introduced BTB3D, a framework that advances vision-language modeling for 3D medical imaging. It unifies 2D and 3D processing through a causal convolutional encoder-decoder and compact volumetric tokenization. Our three-stage training scales learning from local patterns to full-volume anatomical coherence, addressing memory and resolution bottlenecks. BTB3D achieves state-of-the-art results in both radiology report generation and text-conditioned 3D CT synthesis. Notably, the 8$\times$8$\times$8 variant excels in fine-grained tasks such as text-to-CT synthesis, while the 16$\times$16$\times$8 variant is better suited for memory-constrained settings and high-level tasks like report generation. We believe BTB3D is a significant step toward scalable and clinically meaningful vision-language modeling in 3D medical imaging, and we expect its open-source release to catalyze further research.

\begin{ack}

High-performance computing resources were provided by the Erlangen National High Performance Computing Center (NHR@FAU) at Friedrich-Alexander-Universität Erlangen-Nürnberg (FAU), under the NHR projects b143dc and b180dc. NHR is funded by federal and Bavarian state authorities, and NHR@FAU hardware is partially funded by the German Research Foundation (DFG) – 440719683. Additional support was received by the ERC - project MIA-NORMAL 101083647, DFG 512819079, and by the state of Bavaria (HTA). The authors also express sincere gratitude to the Helmut-Horten Foundation for their generous support, which made this work possible.

\end{ack}

\bibliographystyle{unsrt}  
\bibliography{references}

\newpage
\section*{NeurIPS Paper Checklist}

\begin{enumerate}

\item {\bf Claims}
    \item[] Question: Do the main claims made in the abstract and introduction accurately reflect the paper's contributions and scope?
    \item[] Answer: \answerYes{}{} 
    \item[] Justification: The abstract and introduction clearly state two main claims: (1) BTB3D achieves state-of-the-art performance on both radiology report generation and text-conditional 3D CT synthesis, and (2) its improvements stem from a unified causal encoder-decoder architecture and a three-stage training strategy. These claims are thoroughly validated in Section~4 through reconstruction analysis (Section~4.1), report generation benchmarks (Section~4.2), and CT synthesis metrics (Section~4.3). Quantitative improvements (such as a 40\% F1 gain over CT-CHAT for report generation and a 76.5\% FID reduction over GenerateCT) directly support the claims. The paper also discusses architectural trade-offs and generalization, aligning the claims with the demonstrated scope and limitations.

\item {\bf Limitations}
    \item[] Question: Does the paper discuss the limitations of the work performed by the authors?
    \item[] Answer: \answerYes{}{} 
    \item[] Justification: The paper discusses several limitations throughout and includes a dedicated section in the Appendix. Specifically, we acknowledge that BTB3D is evaluated only on chest CTs due to the lack of large-scale paired datasets for other anatomies and modalities. Also, our method requires substantial computational resources for training, which may limit accessibility. Finally, although we evaluate performance using both clinical and generative metrics, further validation through expert reader studies and prospective clinical trials is necessary to assess real-world safety and efficacy.

\item {\bf Theory assumptions and proofs}
    \item[] Question: For each theoretical result, does the paper provide the full set of assumptions and a complete (and correct) proof?
    \item[] Answer: \answerNA{} 
    \item[] Justification: The paper is focused on an empirical vision-language modeling framework for 3D medical imaging and does not contain theoretical results, assumptions, or proofs.

    \item {\bf Experimental result reproducibility}
    \item[] Question: Does the paper fully disclose all the information needed to reproduce the main experimental results of the paper to the extent that it affects the main claims and/or conclusions of the paper (regardless of whether the code and data are provided or not)?
    \item[] Answer: \answerYes{} 
    \item[] Justification: We provide detailed methodology in Section~3 and implementation details in Section~4. Our experiments use the publicly available CT-RATE dataset for training and in-domain evaluation, and RadChestCT for out-of-distribution validation. We release all training and validation scripts, along with pretrained model weights, to ensure full reproducibility of the results and support verification of all main claims.

\item {\bf Open access to data and code}
    \item[] Question: Does the paper provide open access to the data and code, with sufficient instructions to faithfully reproduce the main experimental results, as described in supplemental material?
    \item[] Answer: \answerYes{} 
    \item[] Justification: We will open-source our codebase, including all training, validation, and evaluation scripts. The repository will contain detailed instructions on setting up the environment, accessing and preprocessing the CT-RATE and RadChestCT datasets, and reproducing all experiments reported in the paper. Pretrained weights for BTB3D and the generation models will also be provided to ensure faithful replication of our results.

\item {\bf Experimental setting/details}
    \item[] Question: Does the paper specify all the training and test details (e.g., data splits, hyperparameters, how they were chosen, type of optimizer, etc.) necessary to understand the results?
    \item[] Answer: \answerYes{} 
    \item[] Justification: All training and evaluation details are thoroughly described in Section~4, including dataset splits, preprocessing steps, model configurations, optimizer settings, batch sizes, learning rates, training schedules, and hardware setup. We also specify stage-wise training parameters and evaluation protocols for each task, ensuring the experimental setup is fully transparent and reproducible. 

\item {\bf Experiment statistical significance}
    \item[] Question: Does the paper report error bars suitably and correctly defined or other appropriate information about the statistical significance of the experiments?
    \item[] Answer: \answerNo{} 
    \item[] Justification: Due to the substantial computational cost of training large 3D models on high-resolution CT volumes, we did not conduct repeated runs or report statistical significance metrics such as error bars or confidence intervals. Our focus was on demonstrating consistent performance gains across multiple benchmarks using standardized evaluation protocols. We acknowledge this as a limitation and will consider statistical analysis in future work.

\item {\bf Experiments compute resources}
    \item[] Question: For each experiment, does the paper provide sufficient information on the computer resources (type of compute workers, memory, time of execution) needed to reproduce the experiments?
    \item[] Answer: \answerYes{} 
    \item[] Justification: We detail the hardware and software setup in Section~4. Batch sizes, memory constraints, and stage-wise training configurations are specified, allowing estimation of training time and resource requirements. This enables reproducibility and helps assess the computational cost of our method.
    
\item {\bf Code of ethics}
    \item[] Question: Does the research conducted in the paper conform, in every respect, with the NeurIPS Code of Ethics \url{https://neurips.cc/public/EthicsGuidelines}?
    \item[] Answer: \answerYes{} 
    \item[] Justification: We have carefully reviewed the NeurIPS Code of Ethics and ensured that all aspects of our research (including data usage, model development, evaluation, and reporting) fully comply with its principles. The datasets used are publicly available, de-identified, and ethically sourced, and we maintain transparency and reproducibility throughout the work.

\item {\bf Broader impacts}
    \item[] Question: Does the paper discuss both potential positive societal impacts and negative societal impacts of the work performed?
    \item[] Answer: \answerYes{} 
    \item[] Justification: We provide a discussion of both potential positive and negative societal impacts in the Appendix. On the positive side, BTB3D has the potential to improve diagnostic accuracy, reduce radiologist workload, and enable better access to training tools through high-quality synthetic CT data. On the negative side, we acknowledge potential risks such as misuse of generative models for medical fraud or reconstruction of sensitive information from improperly anonymized data. We emphasize the importance of proper anonymization and include cautionary notes regarding ethical deployment and data governance.
    
\item {\bf Safeguards}
    \item[] Question: Does the paper describe safeguards that have been put in place for responsible release of data or models that have a high risk for misuse (e.g., pretrained language models, image generators, or scraped datasets)?
    \item[] Answer: \answerYes{} 
    \item[] Justification: While we rely solely on publicly available datasets, we acknowledge the potential misuse of generative models. To mitigate this, access to our pretrained models is gated: researchers must apply for access to the model weights, and requests are manually reviewed to ensure responsible use. This safeguard helps prevent fraudulent or unethical deployment of our generation models.

\item {\bf Licenses for existing assets}
    \item[] Question: Are the creators or original owners of assets (e.g., code, data, models), used in the paper, properly credited and are the license and terms of use explicitly mentioned and properly respected?
    \item[] Answer: \answerYes{} 
    \item[] Justification: All external assets used are publicly available datasets, and we comply with their respective licenses (We cite the original papers and follow license terms). Specifically:
    \begin{itemize}
        \item \textbf{CT-RATE dataset:} Released under the CC BY-NC-SA 4.0 license.
        \item \textbf{RadChestCT dataset:} Released under the CC BY-NC-ND 4.0 license.
    \end{itemize}


\item {\bf New assets}
    \item[] Question: Are new assets introduced in the paper well documented and is the documentation provided alongside the assets?
    \item[] Answer: \answerYes{} 
    \item[] Justification: We will release all new assets (including code, pretrained models, and training scripts) under the CC BY 4.0 license. The GitHub repository includes comprehensive documentation covering usage instructions, training configurations, dependencies, and licensing terms to facilitate reproducibility and reuse.

\item {\bf Crowdsourcing and research with human subjects}
    \item[] Question: For crowdsourcing experiments and research with human subjects, does the paper include the full text of instructions given to participants and screenshots, if applicable, as well as details about compensation (if any)? 
    \item[] Answer: \answerNA{} 
    \item[] Justification: This work does not involve crowdsourcing or research with human subjects.

\item {\bf Institutional review board (IRB) approvals or equivalent for research with human subjects}
    \item[] Question: Does the paper describe potential risks incurred by study participants, whether such risks were disclosed to the subjects, and whether Institutional Review Board (IRB) approvals (or an equivalent approval/review based on the requirements of your country or institution) were obtained?
    \item[] Answer: \answerNA{} 
    \item[] Justification: This work does not involve human subjects and therefore does not require IRB.

\item {\bf Declaration of LLM usage}
    \item[] Question: Does the paper describe the usage of LLMs if it is an important, original, or non-standard component of the core methods in this research? Note that if the LLM is used only for writing, editing, or formatting purposes and does not impact the core methodology, scientific rigorousness, or originality of the research, declaration is not required.
    \item[] Answer: \answerNA{} 
    \item[] Justification: LLMs were used only for writing and editing assistance, not as part of the core methodology.

\end{enumerate}

\appendix
\newpage
\section{Technical Appendices and Supplementary Material}
In this supplementary document, we provide additional insights and supporting materials for our main paper. We begin by outlining the key limitations of our work and discussing the broader societal and clinical implications of the BTB3D framework, particularly with regard to its capabilities in radiology report generation and text-conditioned 3D CT volume synthesis. We then present further experimental details that expand upon the methodology described in the main text. Additional qualitative and quantitative results are also included to further validate our findings. All code, pretrained model weights, and instructions to reproduce our experiments will be made openly available in our GitHub repository to promote transparency, reproducibility, and further research in the field.

\subsection{Limitations and Broader Impacts}

\paragraph{Limitations.}
While our BTB3D framework significantly improves tokenization and decoding for 3D medical vision-language modeling (especially for 3D chest CT scans), several limitations remain. First, our current framework does not explicitly model clinical reasoning or uncertainty, both of which are crucial for real-world deployment in clinical settings. Incorporating modules for uncertainty estimation or causal inference remains an open challenge for future studies.

Second, the scope of our experiments is limited to 3D chest CT scans, primarily due to the lack of large-scale, publicly available paired datasets (with reports) for other anatomical regions or modalities (\emph{e.g.}, MRI, PET). While BTB3D is designed to generalize across 2D and 3D inputs, we have not yet validated its transferability to other clinical domains. Future efforts should extend the architecture to whole-body imaging and explore zero-shot or few-shot generalization across organs.

Third, although BTB3D supports both 2D and 3D training modes, we used only 2D slices extracted from 3D volumes in CT-RATE to remain consistent with baselines and ensure fair comparisons. As such, we have not demonstrated its transfer capabilities (\emph{e.g.}, transfer from pretrained 2D models or joint training with 2D paired datasets), which limits the evaluation of BTB3D’s capabilities.

Fourth, statistical analysis is limited due to the high computational cost of 3D training. We did not perform repeated runs or report confidence intervals. Thus, while our results demonstrate strong and consistent performance across multiple benchmarks, statistical significance remains to be validated.

Lastly, although we evaluate both clinical and generative metrics, BTB3D has not yet been assessed in real clinical workflows. Validation through expert reader studies or prospective trials is necessary to fully establish its utility, reliability, and safety in decision-making environments.

\paragraph{Broader impacts.}
BTB3D offers promising benefits for both clinical and research communities. High-quality radiology report generation from 3D CT scans can reduce reporting delays, alleviate radiologist burnout, and improve documentation consistency, particularly in high-volume settings such as emergency departments or large-scale screening programs. Moreover, BTB3D’s ability to synthesize realistic, anatomically plausible CT scans from text opens new avenues for training, simulation, and rare disease modeling. In educational settings, synthetic CT scans can be used to create diverse training sets for radiology students and enhance learning outcomes through exposure to rare or atypical cases. In research, BTB3D may support data augmentation, helping mitigate class imbalance in supervised learning tasks and enabling pretraining in low-resource environments.

However, the generative capabilities of the BTB3D framework also present potential risks. Synthetic scans, if not properly labeled or constrained, may be misused in regulatory or insurance contexts, or exploited for fraudulent purposes. Additionally, there is a non-trivial risk of inadvertently replicating identifiable patient anatomy, even when training data is anonymized. While our work uses only publicly available, fully de-identified datasets (\emph{e.g.}, CT-RATE), developers must ensure strict compliance with data governance and anonymization protocols when deploying similar models. 

Finally, the substantial computational cost of training such models may exacerbate disparities in access to advanced medical AI. Ensuring open-source availability and supporting low-resource inference are critical to democratizing BTB3D’s benefits. Continued oversight and interdisciplinary dialogue will be essential to ensuring the responsible deployment of generative models in healthcare.

\begin{figure}[h]
  \centering
  \includegraphics[width=\textwidth]{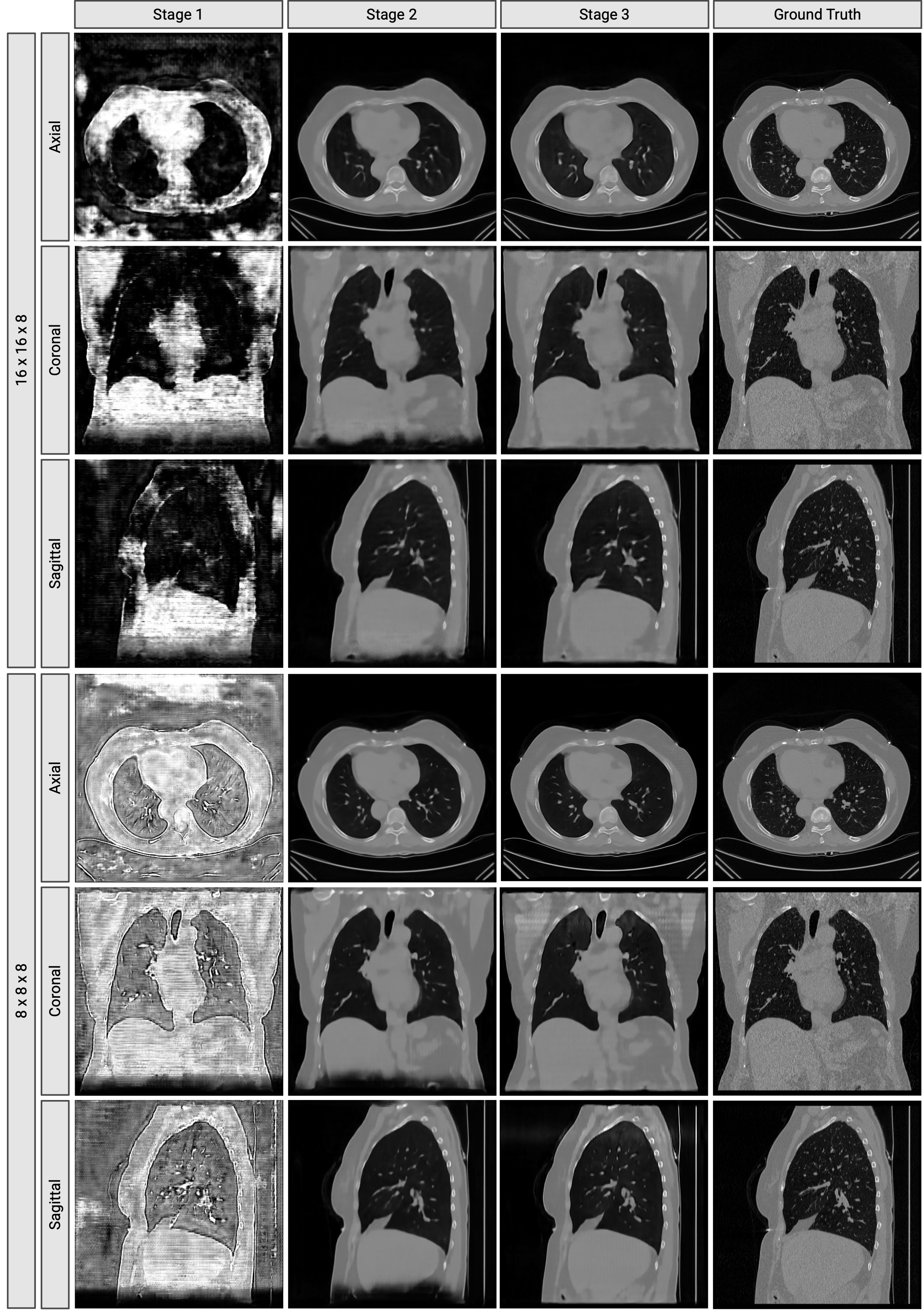}
  
   \caption{Qualitative reconstruction results across axial, coronal, and sagittal planes for two BTB3D variants: 16$\times$16$\times$8 (top) and 8$\times$8$\times$8 (bottom). The figure shows outputs after Stage 1 (short-volume training), Stage 2 (overlapping-window training), and Stage 3 (decoder refinement), compared to the ground truth. The progressive improvements highlight the effectiveness of our three-stage training strategy. Stage 2 yields the largest gain in anatomical fidelity and inter-slice consistency, while Stage 3 further sharpens structural details such as lung fissures and vascular boundaries.}

   \label{fig:3stage_ex}
\end{figure}

\newpage

\subsection{Additional Experimental Details}

\paragraph{Three-stage training performance.}

To further evaluate the impact of our progressive training strategy, we qualitatively assess the reconstruction performance of BTB3D across the three stages of training in Figure~\ref{fig:3stage_ex}. The figure presents axial, coronal, and sagittal slices reconstructed by our two model variants (16$\times$16$\times$8 and 8$\times$8$\times$8) at each stage and compares them to the ground truth.

In \emph{Stage 1}, the model is trained solely on small, non-overlapping 3D volumes and 2D slices. This initialization allows the model to learn basic volumetric structure and inter-slice continuity but suffers from block artifacts and poor spatial coherence, especially in regions with high anatomical complexity. As seen in the leftmost columns of Figure~\ref{fig:3stage_ex}, both variants at this stage produce noisy and blurry reconstructions with limited detail and sharpness. \emph{Stage 2} introduces overlapping windows during training, which significantly improves anatomical consistency across slices and helps reduce the block discontinuities learned in Stage 1. This stage yields the largest qualitative leap in fidelity, with more well-defined boundaries of lung lobes, airways, and soft tissue structures. The improvements are most noticeable in the axial and sagittal views, where inter-slice alignment and smoothness are critical for clinical interpretability. \emph{Stage 3} refines the decoder with high-resolution patches while keeping the encoder part frozen. This final stage enhances local detail, texture realism, and structural sharpness without sacrificing the global consistency achieved in Stage 2. Notably, lung fissures, pleural contours, and fine vascular structures become visibly clearer, indicating that the decoder has learned to reconstruct complex anatomical regions with higher fidelity and resolution.

Comparing the two BTB3D variants, the 8$\times$8$\times$8 model consistently achieves superior anatomical fidelity and inter-slice coherence in reconstruction, as expected due to its lower compression rate and higher capacity. In contrast, the 16$\times$16$\times$8 variant (while more compressed) proves more effective for language-driven tasks such as report generation, where coarse global structure suffices and memory efficiency is critical. Our three-stage training pipeline plays a pivotal role for both models: it first captures global structure, then progressively refines local detail, enabling accurate and high-resolution 3D reconstructions from compact tokens. This staged optimization bridges the gap between token efficiency and clinical utility, facilitating both high-fidelity text-conditional CT generation and precise report generation, underscoring BTB3D’s versatility in multimodal 3D medical image understanding.

\begin{table}[b]
\caption{Abnormality-based F1 scores on CT-RATE along with the prevalence ratio of each abnormality in the test set. BTB3D consistently achieves the highest clinical accuracy across most categories, with the higher-compression variant (16$\times$16$\times$8) demonstrating superior overall performance.}

\label{tab:abnormality_f1}
\centering
\renewcommand{\arraystretch}{1.1}
\begin{tabular}{lcccccc}
\toprule
\textbf{Abnormality} & \textbf{Ratio} & \textbf{Ours-16} & \textbf{Ours-8} & \textbf{CT-CHAT} & \textbf{Merlin} & \textbf{CT2Rep} \\
\midrule
Medical material & 0.103 & 0.142 & 0.120 & 0.006 & 0.057 & 0.000 \\
Arterial wall calcification & 0.285 & 0.414 & 0.273 & 0.451 & 0.262 & 0.322 \\
Cardiomegaly & 0.107 & 0.305 & 0.207 & 0.123 & 0.176 & 0.013 \\
Pericardial effusion & 0.074 & 0.095 & 0.095 & 0.009 & 0.060 & 0.000 \\
Coronary artery wall calc. & 0.252 & 0.403 & 0.260 & 0.412 & 0.235 & 0.335 \\
Hiatal hernia & 0.137 & 0.164 & 0.118 & 0.207 & 0.110 & 0.074 \\
Lymphadenopathy & 0.260 & 0.358 & 0.209 & 0.069 & 0.227 & 0.013 \\
Emphysema & 0.197 & 0.196 & 0.155 & 0.391 & 0.216 & 0.198 \\
Atelectasis & 0.235 & 0.269 & 0.242 & 0.341 & 0.199 & 0.323 \\
Lung nodule & 0.448 & 0.427 & 0.397 & 0.443 & 0.290 & 0.029 \\
Lung opacity & 0.390 & 0.408 & 0.382 & 0.266 & 0.312 & 0.557 \\
Pulmonary fibrotic sequela & 0.273 & 0.318 & 0.211 & 0.069 & 0.117 & 0.104 \\
Pleural effusion & 0.124 & 0.308 & 0.199 & 0.173 & 0.183 & 0.341 \\
Mosaic attenuation pattern & 0.083 & 0.183 & 0.094 & 0.064 & 0.076 & 0.198 \\
Peribronchial thickening & 0.117 & 0.125 & 0.043 & 0.000 & 0.054 & 0.099 \\
Consolidation & 0.191 & 0.259 & 0.185 & 0.120 & 0.174 & 0.236 \\
Bronchiectasis & 0.109 & 0.126 & 0.094 & 0.091 & 0.075 & 0.013 \\
Interlobular septal thick. & 0.082 & 0.135 & 0.087 & 0.075 & 0.065 & 0.032 \\
\midrule
\textbf{Mean} & \textbf{0.193} & \textbf{0.258} & \textbf{0.187} & \textbf{0.184} & \textbf{0.160} & \textbf{0.160} \\
\bottomrule
\end{tabular}
\end{table}

\paragraph{Radiology report generation from 3D chest CT.}
We comprehensively evaluate BTB3D's capability in generating accurate and clinically coherent radiology reports from volumetric chest CT scans. As described in the main paper, each CT volume is first compressed into a compact sequence of frequency-aware 3D tokens by our encoder, and these tokens are then passed to a pretrained LLM (LLaMA 3.1 8B) for report generation via a linear projection layer. To ensure fairness, we use the official weights for CT2Rep and CT-CHAT, both trained on the same CT-RATE dataset. Since Merlin was originally trained on a private dataset and neither its weights nor training data are publicly available, we retrained Merlin on CT-RATE using its official codebase and default hyperparameters.

Our evaluation includes both internal (CT-RATE test set) and external (RAD-ChestCT) benchmarks. The results in Table~\ref{tab:abnormality_f1} show that BTB3D (particularly the 16$\times$16$\times$8 variant) outperforms prior methods in average abnormality level F1 scores, achieving higher clinical precision in most of the findings. This trend holds in Table~\ref{tab:abnormality_f1_external}, where our BTB3D framework demonstrates strong out-of-distribution generalization, with the 16$\times$16$\times$8 variant yielding a 46\% relative F1 improvement over CT-CHAT (the previous state-of-the-art method). This highlights the robustness of our tokenization and training pipeline, even when evaluated on unseen institutional distributions.

\begin{figure}[t!]
  \centering
  \includegraphics[width=\textwidth]{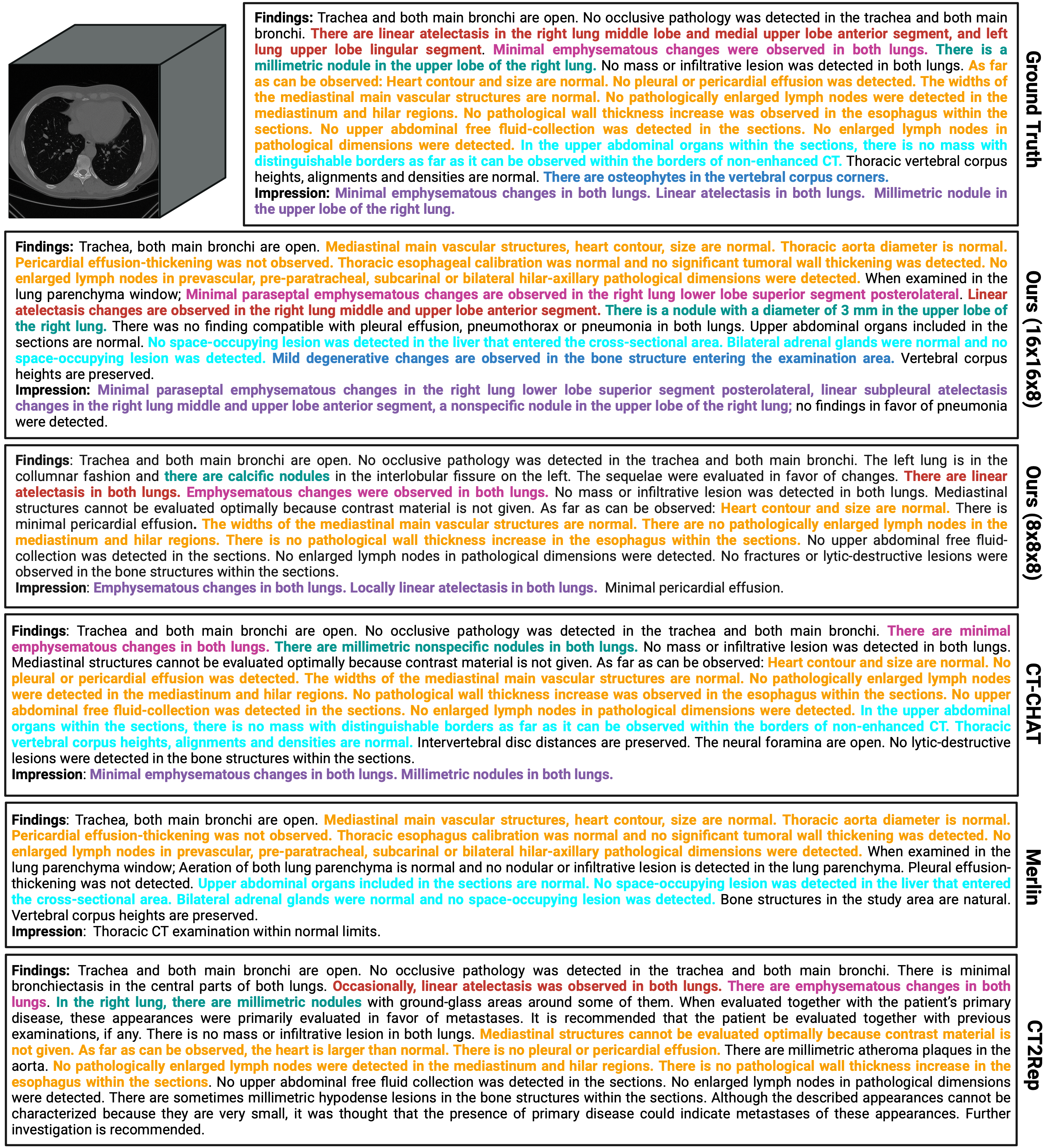}
  \caption{Example of radiology report generation for the same 3D chest CT scan using our BTB3D method (both 16$\times$16$\times$8 and 8$\times$8$\times$8 variants) compared to baseline models (CT-CHAT, Merlin, CT2Rep) and the ground truth report. Key phrases from the ground truth are highlighted and matched across model outputs using consistent colors to indicate alignment. Our BTB3D framework, especially the higher compression rate variant, produces more detailed, clinically relevant, and accurate radiology reports, showing superior coverage of anatomical structures and abnormalities.}
  \label{fig:rep_ex}
\end{figure}

Figure~\ref{fig:rep_ex} offers a qualitative comparison across models, illustrating report generation for the same CT scan. BTB3D’s reports more faithfully reproduce key clinical details from the ground truth, particularly with the higher compression (16$\times$16$\times$8) variant, supporting the notion that coarser representations, while less suitable for pixel-level synthesis, may better capture global semantics for language modeling. Meanwhile, the 8$\times$8$\times$8 variant provides more spatially detailed reconstructions but slightly underperforms on text generation metrics, suggesting a trade-off between compression depth and semantic abstraction. In all comparisons, BTB3D demonstrates a compelling advantage by combining precise volumetric tokenization with a scalable training strategy, ultimately allowing both fine-grained anatomical reconstruction and clinically relevant report generation.

\begin{table}[t]
\caption{Abnormality-wise F1 scores on the RAD-ChestCT dataset (external test set), along with the prevalence ratio of each abnormality. BTB3D demonstrates strong generalization performance, particularly with the 16$\times$16$\times$8 variant, achieving the highest F1 score across most categories.}
\label{tab:abnormality_f1_external}
\centering
\renewcommand{\arraystretch}{1.1}
\begin{tabular}{lcccccc}
\toprule
\textbf{Abnormality} & \textbf{Ratio} & \textbf{Ours-16} & \textbf{Ours-8} & \textbf{CT-CHAT} & \textbf{Merlin} & \textbf{CT2Rep} \\
\midrule
Medical material & 0.327 & 0.235 & 0.154 & 0.000 & 0.107 & 0.000 \\
Calcification & 0.706 & 0.671 & 0.406 & 0.567 & 0.359 & 0.434 \\
Cardiomegaly & 0.109 & 0.187 & 0.151 & 0.181 & 0.031 & 0.010 \\
Pericardial effusion & 0.155 & 0.193 & 0.101 & 0.007 & 0.047 & 0.089 \\
Hiatal hernia & 0.117 & 0.149 & 0.087 & 0.149 & 0.136 & 0.173 \\
Lymphadenopathy & 0.165 & 0.252 & 0.227 & 0.121 & 0.176 & 0.000 \\
Emphysema & 0.273 & 0.201 & 0.172 & 0.412 & 0.242 & 0.142 \\
Atelectasis & 0.298 & 0.350 & 0.265 & 0.387 & 0.194 & 0.153 \\
Lung nodule & 0.802 & 0.424 & 0.408 & 0.721 & 0.495 & 0.068 \\
Lung opacity & 0.539 & 0.542 & 0.389 & 0.140 & 0.394 & 0.560 \\
Pulmonary fibrotic sequela & 0.132 & 0.204 & 0.139 & 0.026 & 0.174 & 0.157 \\
Pleural effusion & 0.200 & 0.290 & 0.199 & 0.043 & 0.135 & 0.032 \\
Peribronchial thickening & 0.085 & 0.117 & 0.064 & 0.000 & 0.095 & 0.039 \\
Consolidation & 0.139 & 0.217 & 0.145 & 0.108 & 0.149 & 0.264 \\
Bronchiectasis & 0.158 & 0.128 & 0.084 & 0.034 & 0.122 & 0.000 \\
Interlobular septal thick. & 0.069 & 0.091 & 0.075 & 0.008 & 0.057 & 0.000 \\
\midrule
\textbf{Mean} & \textbf{0.267} & \textbf{0.266} & \textbf{0.192} & \textbf{0.182} & \textbf{0.182} & \textbf{0.133} \\
\bottomrule
\end{tabular}
\end{table}

\newpage
\paragraph{Text-conditional 3D chest CT generation.}

\begin{figure}[t]
  \centering
  \includegraphics[width=0.9\textwidth]{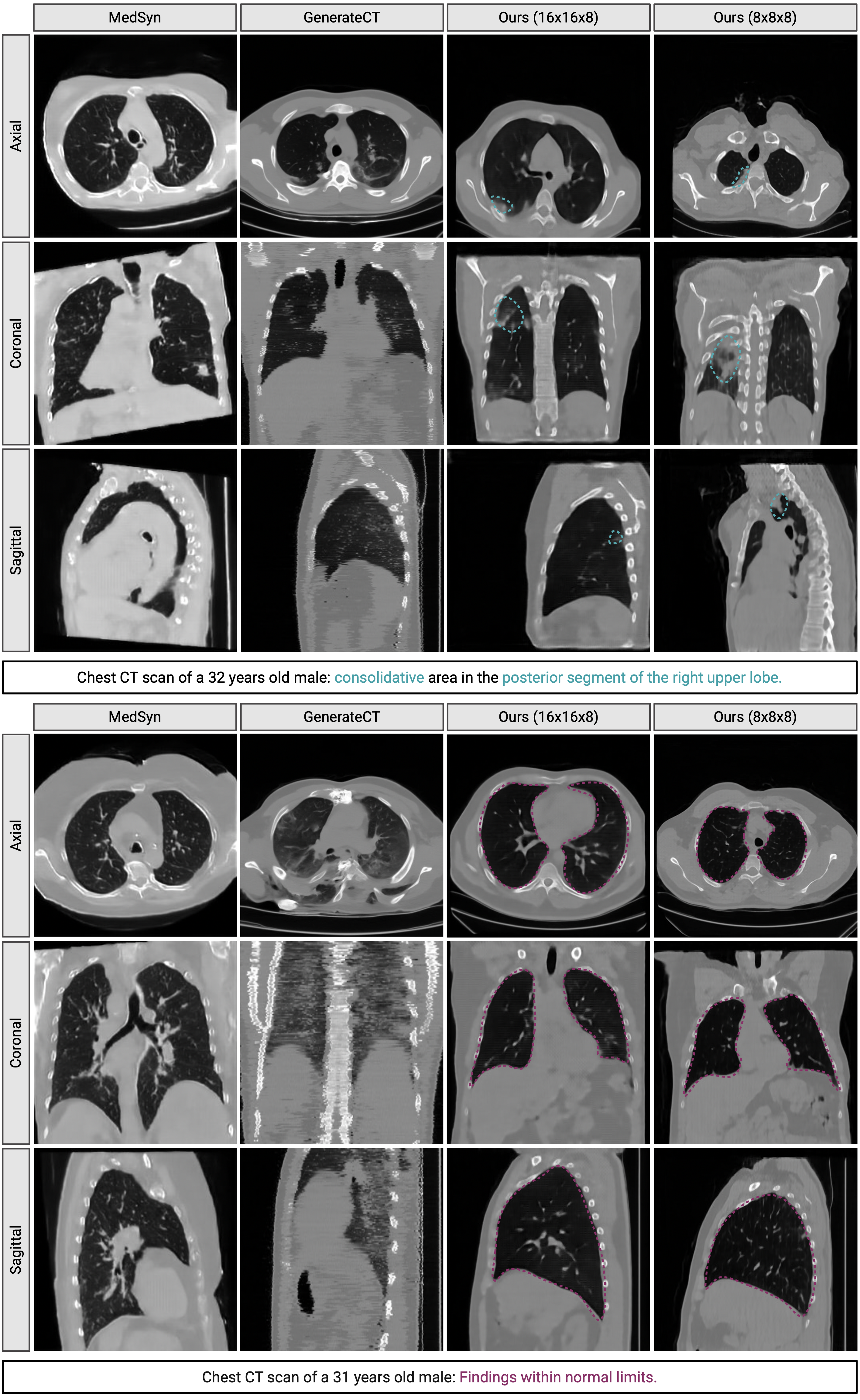}
\caption{Text-conditioned CT generation results for two clinical prompts using MedSyn, GenerateCT, and our BTB3D models. For each case, we show one representative slice per anatomical plane. The lower-compression variant of BTB3D produces the most consistent volumes, demonstrating superior alignment with the prompt. Prompts and corresponding anatomical regions are highlighted using color overlays. Ground-truth volumes are omitted, following standard practice in generative tasks.}

  \label{ct_generation_fig_ex}
\end{figure}

To evaluate the generative capabilities of BTB3D, we benchmark its performance on synthesizing realistic and anatomically coherent 3D chest CT volumes from free-text clinical prompts. As shown in Figure~\ref{ct_generation_fig_ex}, BTB3D generates sharper, more anatomically faithful volumes compared to MedSyn and GenerateCT, with improved inter-slice consistency and alignment to prompt semantics. For fair comparison, we use the official pretrained weights for both MedSyn and GenerateCT, which were trained on the same modality (3D chest CT), ensuring that differences in performance stem from architectural and training innovations.

Quantitative results reported in the main paper (Table 4) show that our BTB3D framework with the lower compression variant (8$\times$8$\times$8) achieves the best overall performance across all generative metrics. Specifically, it reduces the mean Fréchet Inception Distance (FID) from 9.51 (GenerateCT) and 12.59 (MedSyn) to just 2.24, a 76.5\% improvement, demonstrating superior fidelity. We compute FID using the \texttt{FIDMetric} from the MONAI library~\cite{cardoso2022monai}, leveraging a RadImageNet-pretrained ResNet50 backbone~\cite{radimagenet}, which is better suited for grayscale radiology images than traditional Inception networks. Following standard practice established by GenerateCT and MAISI~\cite{guo2025maisi}, FID is calculated on the central 40\% of slices (in each anatomical plane) across 100 randomly selected volumes per method, reducing boundary noise and focusing on clinically relevant regions.

In terms of temporal and anatomical realism, Fréchet Video Distance (FVD) is computed using both CT-Net (specialized for 3D chest CTs)~\cite{draelos2021machine} and I3D (trained on RGB videos). BTB3D again significantly outperforms prior work, halving the FVD compared to GenerateCT. To assess semantic alignment between text prompts and generated volumes, we compute CLIP scores using the \texttt{CLIPScore} implementation from Torchmetrics. We follow GenerateCT’s protocol: axial slices are resized to $224 \times 224$ and converted to pseudo-RGB by repeating the single intensity channel. Using \texttt{clip-vit-base-patch16}, we observe that BTB3D achieves the highest text-image alignment score (24.27), suggesting it captures fine semantic cues better than prior methods.

Interestingly, we note a compression-quality trade-off: while the 8$\times$8$\times$8 variant excels in fine-grained reconstruction and text-conditional volume synthesis, the more compact 16$\times$16$\times$8 variant still surpasses existing baselines and may be preferable in memory-constrained or latency-sensitive settings. Together, these results confirm that BTB3D’s volumetric tokenization and three-stage training pipeline offer a significant leap forward in text-conditional 3D medical image generation, bridging the gap between semantic understanding and pixel-level anatomical coherence.

\end{document}